\documentclass[12pt]{l4dc2022}

\title[CEM-GD Planner for Model-Based RL]{CEM-GD: Cross-Entropy Method with Gradient Descent Planner \\ for Model-Based Reinforcement Learning}
\usepackage{times}
\usepackage[english]{babel}
\usepackage{color}
\usepackage[noend,ruled]{algorithm2e}
\usepackage{caption}
\usepackage{float}
\usepackage{wrapfig}

\usepackage{xspace}
\usepackage{enumitem}

\newcommand{\alg}{\textsc{\small{CEM-GD}}\xspace}

\graphicspath{ {./figures/} }

\author{%
 \Name{Kevin Huang} \Email{khuang@caltech.edu}\\
 \Name{Sahin Lale} \Email{alale@caltech.edu}\\
 \Name{Ugo Rosolia} \Email{ugo.rosolia@gmail.com}\\
 \Name{Yuanyuan Shi} \Email{yyshi@eng.ucsd.edu}\\
 \Name{Anima Anandkumar} \Email{anima@caltech.edu}
}

\begin{document}

\maketitle

\begin{abstract}
Current state-of-the-art model-based reinforcement learning algorithms use trajectory sampling methods, such as the Cross-Entropy Method (CEM), for planning in continuous control settings. These zeroth-order optimizers require sampling a large number of trajectory rollouts to select an optimal action, which scales poorly for large prediction horizons or high dimensional action spaces. First-order methods that use the gradients of the rewards with respect to the actions as an update can mitigate this issue, but suffer from local optima due to the non-convex optimization landscape. To overcome these issues and achieve the best of both worlds, we propose a novel planner, Cross-Entropy Method with Gradient Descent (\alg), that combines first-order methods with CEM. At the beginning of execution, \alg uses CEM to sample a significant amount of trajectory rollouts to explore the optimization landscape and avoid poor local minima. It then uses the top trajectories as initialization for gradient descent and applies gradient updates to each of these trajectories to find the optimal action sequence. At each subsequent time step, however, \alg samples much \textit{fewer} trajectories from CEM before applying gradient updates. We show that as the dimensionality of the planning problem increases, \alg maintains desirable performance with a constant small number of samples by using the gradient information, while avoiding local optima using initially well-sampled trajectories. Furthermore, \alg achieves better performance than CEM on a variety of continuous control benchmarks in MuJoCo with $100$x fewer samples per time step, resulting in around $25\%$ less computation time and $10\%$ less memory usage. The implementation of \alg is available at \url{https://github.com/KevinHuang8/CEM-GD}.


\end{abstract}
\begin{keywords}%
  Control and Planning, Cross-Entropy Method, Nonlinear Systems, Model-based RL%
\end{keywords}

\section{Introduction}
Recently, there has been a flurry of exciting deployments and advances in model-based reinforcement learning (MBRL) for challenging control tasks \citep{pets, poplin, Hafner, lale2021model}. These state-of-the-art MBRL methods typically solve a planning problem, where the goal is to find the optimal sequence of actions that maximize the cumulative reward according to a learned dynamics model, and execute actions with model-predictive control (MPC), meaning that replanning occurs at each time step. These planning problems are usually too complex to solve optimally due to highly nonlinear system dynamics, continuous action spaces, and complicated, possibly non-convex, cost functions. Therefore, for such tasks, the standard approach of finding the optimal action sequences is the family of population-based sampling methods such as Random-Sampling Shooting Method (RS)~\citep{nagabandi2018neural}, Model Predictive Path Integral Control (MPPI)~\citep{williams2017model}, and Cross-Entropy Method (CEM)~\citep{rubinstein1997optimization}. 

Fundamentally, these zeroth-order optimizers are variants of CEM, which maintains a sampling distribution to sample rollouts of action sequences~\citep{okada2020variational}. This sampling distribution is updated at each iteration to assign a higher probability near higher reward action sequences. However, CEM, as well as the family of all sampling-based approaches, suffers from scalability for high-dimensional problems \citep{gradcem}. As the dimension of the action space and/or the planning horizon grows, many more samples are needed for CEM-based planners to converge, and the variance in the reward of CEM-based planners increases \citep{gradcem, pets}. 
This effect makes it difficult to use CEM with long planning horizons if the number of samples taken by CEM is fixed \citep{pets}. Scalability is a fundamental drawback of zeroth-order optimizers, and poses a significant challenge to real-time planning when computation time is a constraint, e.g., when parallelization is not possible.

First-order methods are often used effectively on high-dimensional optimization problems. In MBRL, if we have a differentiable dynamics model and objective function, we can directly optimize the action sequences following the gradient direction of the reward function with respect to actions~\citep{Hafner, srinivas2018universal}. However, first-order methods are susceptible to poor local optima~\citep{von1992direct}. Indeed, as we show in our experiments, the performance of a naive approach of selecting a single random initialization point for gradient descent, i.e., a first-order optimization approach, varies wildly across different environments due to the existence of local minima. 

An intriguing idea is to combine zeroth-order and first-order planning methods in order to enjoy the benefits of both simultaneously. To this end, \citet{gradcem} recently proposed to perform gradient updates within CEM itself by interleaving CEM iterations with gradient updates. Even though this method shows improved convergence and performance over CEM in high-dimensional tasks, it performs gradient updates to each sample in the population, which is computationally expensive and thus still has issues with scalability to higher dimensional planning problems.  

\begin{figure}[t]
\centering
\includegraphics[width=1\textwidth]{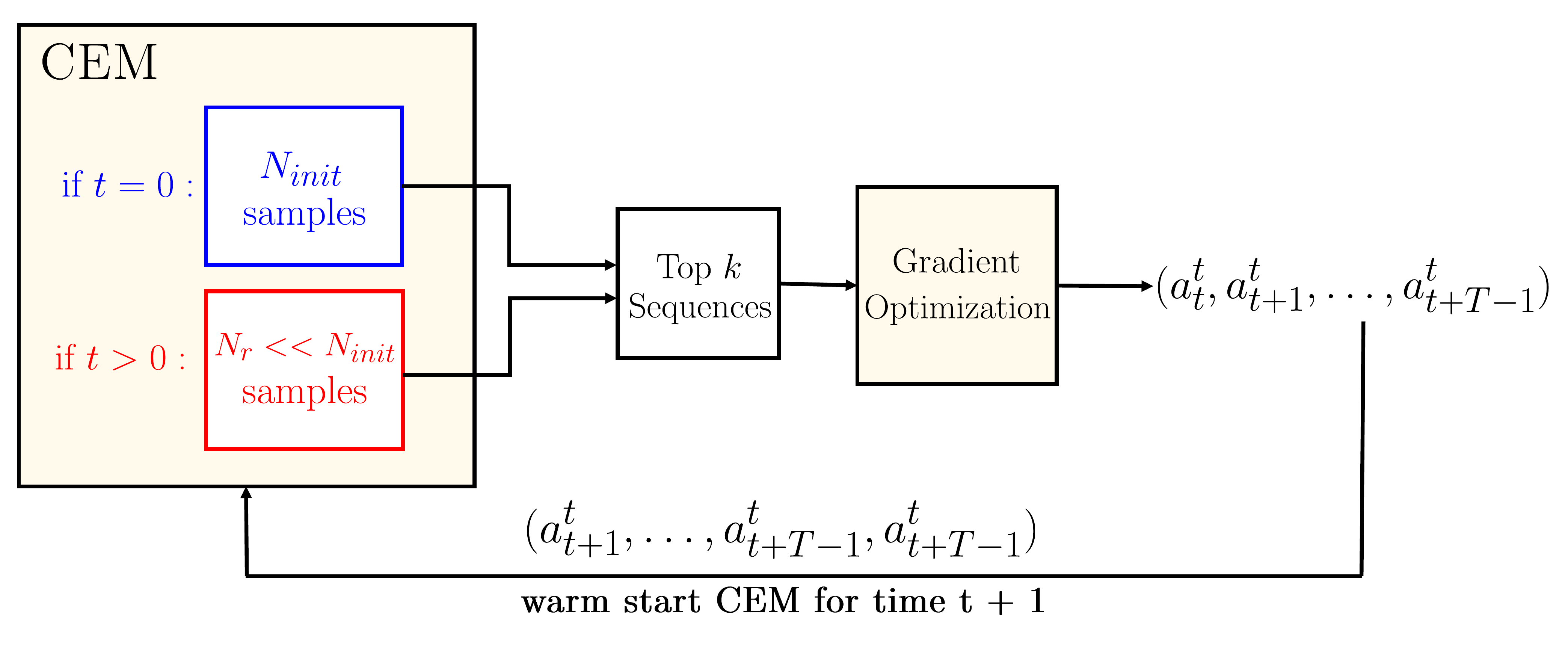}
\caption{\label{fig:overview} Overview of \alg. At time $t = 0$, we take $N_{init}$ samples using CEM with a starting zero-mean distribution. The top $k$ action sequences are taken as initialization points for gradient optimization, with the best final action sequence being the output of the routine. For $t > 0$, we take $N_{r} << N_{init}$ samples, and we initialize the CEM distribution at time $t+1$ with a mean at the time shifted optimal sequence from time $t$, with the last action initialized as the same as the previous one.}
\end{figure}

\subsection*{Our contributions:} 
In this paper, we propose a new \textit{scalable} gradient-based planner named Cross-Entropy Method with Gradient Descent (CEM-GD), that combines zeroth-order optimization, specifically CEM, with first-order optimization. \alg first uses CEM to generate initialization points for optimization. It then selects the top $k$ action sequences with the highest rewards and performs gradient updates on them. After the optimization, \alg uses the highest-reward achieving action sequence for actuation. Finally, \alg warm-starts the mean of the CEM sampling distribution in the next time step as the current optimal action sequence to guide the sampling towards high-reward regions, as shown in Figure \ref{fig:overview}. We show that 
\begin{enumerate}[wide, labelindent=0pt,label=\textbf{\arabic*})]
    \item \alg achieves the best of both worlds and obtains better scalability in sample complexity for high dimensional planning problems than the state-of-the-art planner CEM, while also mitigating the problem of poor local optima that first-order methods can suffer.
    \item On a variety of high dimensional continuous control tasks, \alg achieves better performance than CEM, with 100 times fewer sampled rollouts, yielding $25\%$ less computation time and $10\%$ decrease in memory usage.
\end{enumerate}


The key intuition is that by including the gradient updates, we only need a few samples from CEM, even when the dimensionality is high. 
Thus, as the dimensionality of the problem grows, the standard CEM planner requires sampling more and more rollouts to maintain the same level of performance, whereas our \alg planner requires a roughly constant, small number of rollouts as a function of dimensionality. 
Furthermore, \alg allows us to alleviate the problem of local minima of pure first-order methods, leveraging the stochasticity from the zeroth-order sampling approach. In particular, at the beginning of execution, \alg samples a large number of trajectory rollouts via CEM, to ensure that our first action sequence is not at a poor local minima, 
while in later time steps it only samples a small number of initial trajectories. This hybrid approach proves to be effective in several benchmark control tasks, maintaining or exceeding the performance of state-of-the-art zeroth-order planners while enjoying lower sample-complexity, resulting in less computation time and memory usage.

\section{Methods}

\subsection{Preliminaries: Cross-Entropy Method for Trajectory planning}
In model-based reinforcement learning~\cite{nagabandi2018neural}, a common scheme for action selection is to use model predictive control (MPC). At each time step $t$, 
the planner needs to solve the following finite time optimal control problem, 
\begin{subequations}\label{eq:opt_problem}
    \begin{align}
    \underset{a_t, \dots, a_{t + T - 1} \in \mathcal{A}^T}{\text{argmax}} \quad &\sum_{\tau = t}^{t + T - 1} r(s_{\tau + 1}, a_{\tau}) \\
     \text{s.t. } \quad \quad & s_{\tau + 1} = f(s_{\tau}, a_{\tau}), \quad \text{for } \tau = t, \dots, t + T - 1
\end{align}
\end{subequations}
where $\mathcal{S}$ and $\mathcal{A}$ denote the state and action spaces, $T$ is the planning horizon, and $r: \mathcal S \times \mathcal A \rightarrow \mathbb{R}$ denotes the reward function, which we assume is known. $\mathcal{A}^T$ refers to the space of action sequences of length $T$ that we are optimizing over, i.e. $\mathcal{A} \times \dots \times \mathcal{A}$ for $T$ total times. Define $R(s_{0:T}, a_{0:T-1}) := \sum_{t = 0}^{T - 1} r(s_{t + 1}, a_t)$ as the total reward of the action sequence starting from state $s_0$. The function $f : \mathcal{S} \times \mathcal{A} \rightarrow \mathcal{S}$ represents the system dynamics, which in the context of MBRL, is a differentiable model which can be learned from data.
Let $(a^t_t, a^t_{t + 1}, \dots, a^t_{t + T-1})$ be the solution to the optimal control problem defined in ~\eqref{eq:opt_problem}. The MPC policy selects the first action $a^t_t$ of the optimal control sequence to execute, and replan at the next time step with the new state.

CEM is a zeroth-order method to solve the optimal control problem in Eq.~\eqref{eq:opt_problem}. It maintains a sampling distribution, commonly a Gaussian $\mathcal{N}(\mu, \Sigma)$, over $\mathcal{A}^T$. At each iteration, $n$ action sequences are sampled from this distribution, and the top $k$ sequences with the highest reward will be used to refine the distribution. This process is repeated for $m$ total iterations, for a total of $N = nm$ sampled rollouts. In particular, CEM keeps track of the moving average of the mean and variance of the top $k$ reward action sequences at each iteration. That is, the new mean and variance at iteration $i$ is updated to be the weighted average of the previous mean $\mu^{i - 1}$ and standard deviation $\Sigma^{i - 1}$ and the mean and variance of the top $k$ reward action sequences sampled at iteration $i - 1$. For $i = 1, ..., m$,
\begin{subequations}\label{eq:cem_update}
\begin{align}
\mu^{(i)} &\leftarrow (1 - \alpha)\mu^{(i - 1)} + \alpha \left(\text{mean}(\{a^{(i - 1)}_{t:t + T - 1}\}_{j = 1}^k)\right), \\
\Sigma^{(i)} &\leftarrow (1 - \alpha)\Sigma^{(i - 1)} + \alpha  \left(\text{variance}(\{a^{(i - 1)}_{t:t + T - 1}\}_{j = 1}^k)\right),
\end{align}
\end{subequations}
where $0 < \alpha < 1$, and $1 - \alpha$ represents the discount rate. After $m$ such updates, the highest reward action sequence seen is selected as the optimal action sequence $(a^t_t, a^t_{t + 1}, \dots, a^t_{t + T-1})$, and the MPC policy chooses the first action $a^t_t$ to execute. Recall that CEM takes samples over $\mathcal{A}^T$. Thus, as the dimension of $\mathcal{A}$ grows or as the time horizon $T$ grows, the size of the sample space greatly increases and the number of samples needed to converge to an optimal solution increases \citep{gradcem}. 

\subsection{Leveraging Gradients in Planning}\label{gradient_descent}
When the model $f$ is differentiable, it is possible to leverage first-order methods to solve the control problem from Eq.~\eqref{eq:opt_problem}.
Given an initial sequence $a^{(0)}_{t:t+T-1}$, we can apply the gradient update repeatedly to obtain a (locally) optimal action sequence
\begin{equation}\label{eq:action_update}
    a_{t:t+T-1}^{(i)} \gets a_{t:t+T-1}^{(i - 1)} + \eta \nabla_{a_{t:t+T-1}} R(s_{t+1:t+T}, a_{t:t+T-1}^{(i - 1)})\,,
\end{equation}
where $s_{t+1:t+T}$ is obtained by rolling out the action sequence $a_{t:t+T-1}$ in the model starting from the current state $s_t$. 
To select a step size $\eta$, we employ a backtracking line search~\citep{hauser2007line} at each step. The line search is performed by first setting $\eta$ to a initial value $\eta_{init}$. If $R(s_{t+1:t+T}, a_{t:t+T-1}) + \eta \nabla_{a_{t:t+T-1}} R(s_{t+1:T}, a_{t:t+T-1}) < R(s_{t+1:t+T}, a_{t:t+T-1})$, i.e. if the reward decreases, $\eta$ decreases by a factor of $\rho$, with $\rho < 1$. Otherwise, we take a gradient step to update the action sequence following Eq~\eqref{eq:action_update}. If a sufficiently small $\eta$ is not found in $J$ iterations, the gradient update ends and returns the optimal action sequence $a_{t:t+T-1}^{(J)}$.

To ensure that the action sequence is within the state and action constraint set, we use a projected gradient descent approach. If a step were to take the sequence outside of its constraints, the actions would be projected back to the feasible set. For example, if we have a constraint $\alpha \leq a_i \leq \beta$, we project by clamping the value of $a_i$ to the range $[\alpha, \beta]$.

\subsection{CEM with Gradient Descent Planner (\alg)}
In general, the optimization landscape of the planning problem as described in Equation \eqref{eq:opt_problem} is highly nonconvex, and thus the performance of gradient descent relies heavily upon the initialization point; gradient descent can easily lead to a poor local optima. \alg overcomes this issue by using CEM to select the initial action sequences that gradient descent starts from. 

At $t = 0$  \alg samples $N_{init}$ rollouts using CEM to obtain a good initial trajectory. For time $t > 0$, \alg samples $N_{r}$ rollouts using CEM, with $N_{r} << N_{init}$, with the mean of the sampling distribution warm started with the optimal action sequence at the previous time step. That is, if the optimal action sequence found at time $t - 1$ was $(a^{t - 1}_t, a^{t - 1}_{t + 1}, \dots, a^{t - 1}_{t + T-1})$, we initialize the sampling distribution of CEM for time $t$ with $\mu^{t} = (a^{t - 1}_{t + 1}, \dots, a^{t - 1}_{t + T-1}, a^{t - 1}_{t + T-1})$ and $\Sigma = \mathbf{I}$. Using CEM, $N_{r}$ action sequences are sampled from an initial sampling distribution of $N(\mu^{t}, \Sigma)$, and the top $k$ sequences are chosen as initial starting points for gradient descent. After optimization, the optimized sequence with the highest reward is chosen as the optimal plan. The planning algorithm is shown in Algorithm \ref{alg:main_alg}.

\SetKwComment{Comment}{/* }{ */}
\SetKwInOut{Input}{Inputs}
\SetKw{Break}{break}

\begin{algorithm}[t]
\caption{\alg}\label{alg:main_alg}
\Input{current time step $t$, current state $s_t$}

 Sample $N$ action sequences $\{a_{t:t + T - 1}^{(i)}\}$, $i = 1\dots N$ using CEM with starting mean $\mu$, and variance $\Sigma = \mathbf{I}$. If $t = 0$, $N = N_{init}$ and $\mu = \mathbf{0}$. Otherwise, if $t > 0$, $N = N_{r}$ and $\mu = (a^{t - 1}_t, \dots a^{t - 1}_{t + T - 1}, a^{t - 1}_{t + T - 1})$ \\
Rollout each action sequence using the dynamics model \\
\For{$i = 1$ \KwTo $N$} {
    \For{$\tau = 1$ \KwTo $T$} {
        $s^{(i)}_{t + \tau} \gets f(s^{(i)}_{t + \tau - 1}$, $a^{(i)}_{t + \tau - 1})$\\
    }
}
Take the $k$ action sequences with highest reward $\{a_{t:t + T - 1}^{(i)}\}$, $i = 1 \dots k$ \\
Optimize each action sequence by applying $G$ gradient updates, trying $J$ different step sizes each update. \\
\For{$i = 1$ \KwTo $k$} {
    \For{$j = 1$ \KwTo $G$} {
        \For{$\text{trial} = 1$ \KwTo $J$} {
            $a{'}_{t:t+T-1} \gets a_{t:t+T-1}^{(i)} + \eta \nabla_{a_{t:t+T-1}^{(i)}} R(s_{t+1:t+T}, a_{t:t+T-1}^{(i)})$\\
            \text{Obtain $s'_{t+1:t+T}$ by rolling out $a'_{t:t+T-1}$}\\
            \If {$R(s'_{t+1:t+T}, a'_{t:t+T-1}) > R(s_{t+1:t+T}, a_{t:t+T-1})$} {
                $a_{t:t+T-1}^{(i)} \gets a'_{t:t+T-1}$\\
                \Break
            }
            \Else {
                $\eta \gets \rho \eta$ \\
            }
        }
    }

}

Rollout each action sequence $\{a_{t:t + T - 1}^{(i)}\}$, $i=1\dots k$ and choose the action sequence with the highest reward $(a^t_t, \dots, a^t_{t + T - 1})$  \\

Execute $a^t_t$ with the agent in the true environment
\end{algorithm}



\begin{figure}[t]
	\centering
	\subfigure[]{\label{fig:toyenvfail}\includegraphics[width=0.45\columnwidth]{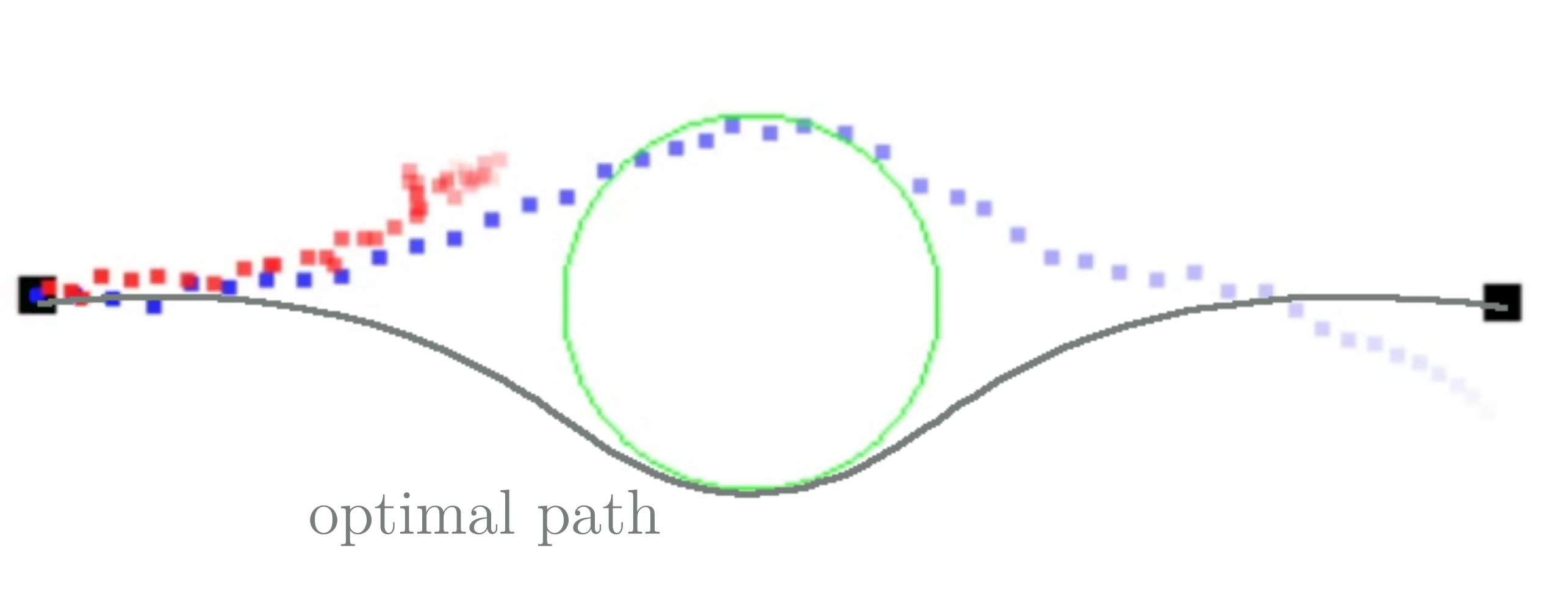}} 
	\subfigure[]{\label{fig:toyenvsuccess}\includegraphics[width=0.45\columnwidth]{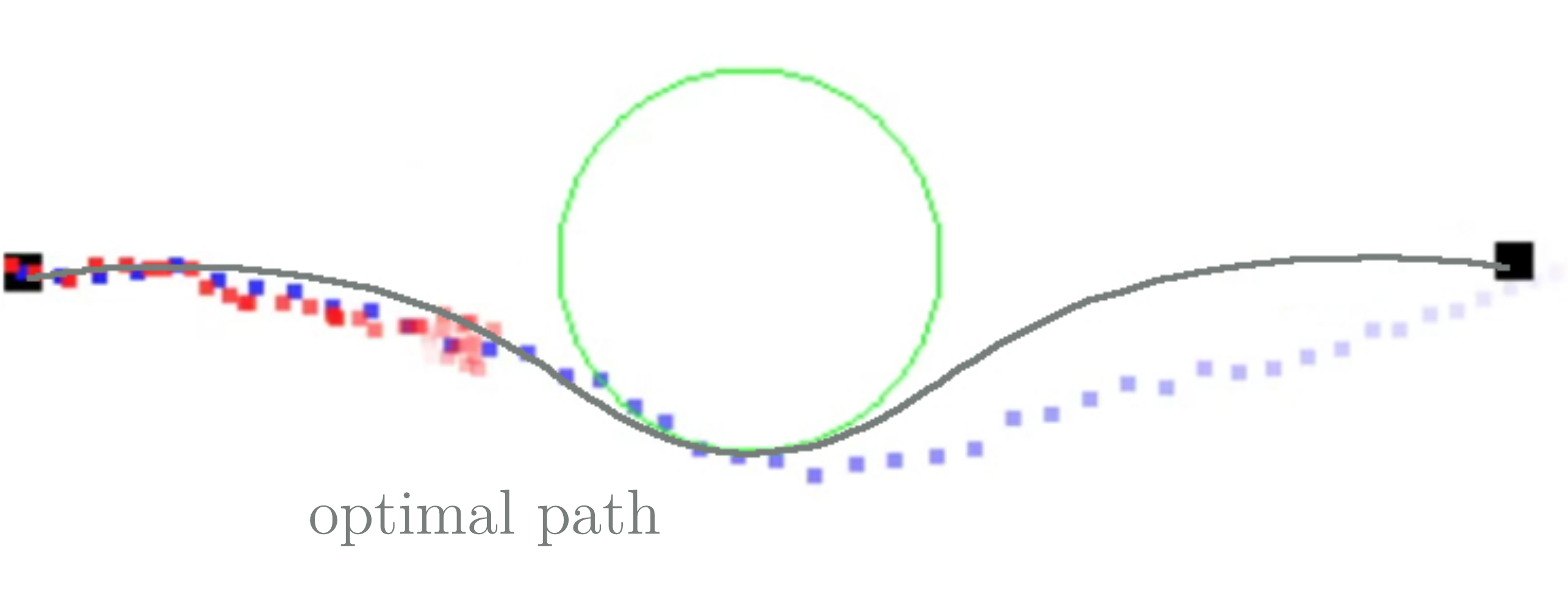}} 
	\subfigure[]{\label{fig:ninit}\includegraphics[width=0.7\columnwidth]{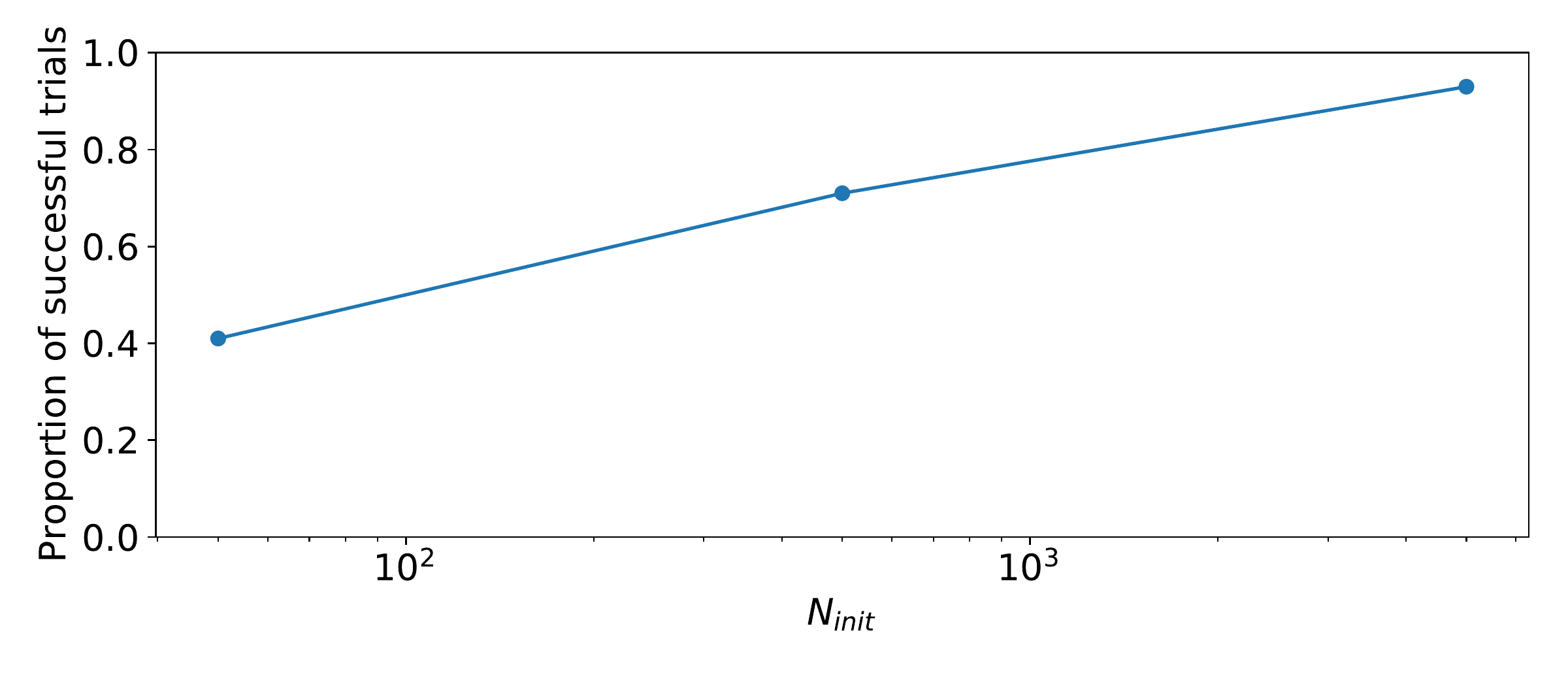}} 
	\caption{(a) An visualization of an suboptimal trajectory. The red trajectory represent the initialization point selected by the gradient planner using $N_{init} = 50$, the blue trajectory represents the final optimized trajectory after applying gradient descent on the red trajectory. (b) An visualization of a successful trajectory, where the red trajectory represents the initialization trajectory of the gradient planner using $N_{init} = 500$, the blue trajectory represents the final optimized trajectory after applying gradient descent on the red trajectory. (c) For various values of $N_{init}$, the proportion of trials in which the agent successfully travels beneath the barrier.}
	\label{fig:localoptima}
\end{figure}

A key observation here is that a large number of samples at the initial time step (i.e., $N_{init}$) can effectively remedy the poor local optima issue. 
Consider a toy environment, shown in Figure \ref{fig:localoptima}, which an agent seeks to move to a goal on a 2D plane. Standing in between the starting location of the agent and the goal is a circular potential barrier, which repels the agent if it enters. The barrier is placed slightly offset from the $y = 0$ line, such that it is shorter to traverse below the barrier than above it. The path above the barrier represents a sub-optimal local optimum, while the path below represents the true optimum.
Running our gradient planner on this toy environment, we show the difference in performance for different levels of $N_{init}$ in Figure \ref{fig:ninit}. 
We see that as $N_{init}$ increases, the planner takes the bottom, global optimum more often. Note that we only vary $N_{init}$, the number of samples at time $t = 0$. At every other time step, we take $N_{r} = 50$ samples.

By including the gradient updates, the CEM-GD planner only needs a small number of samples except for the initial time step, even when the dimensionality is high. Specially, when the problem dimension increases, the standard CEM planner requires sampling more and more rollouts to maintain the same level of performance whereas CEM-GD requires a roughly constant, small number of rollouts. 
In experiments, CEM-GD maintains or exceeds the performance of state-of-the-art CEM planners while enjoying lower sample-complexity, computation time and memory usage.

\section{Experiments}
We demonstrate the efficiency of the proposed CEM-GD algorithm for planning on four benchmark continuous control tasks: the Cartpole Swingup, Half-Cheetah, Pusher, and Humanoid Standup. The latter three are implemented in Mujoco \citep{mujoco}, while Cartpole Swingup is implemented in OpenAI Gym \citep{gym}. We use the state-of-the-art PETS algorithm \cite{pets} as our end-to-end reinforcement learning algorithm, and compare our planner with two baseline planners:
\begin{itemize}
    \item \textbf{Baseline CEM ($N$ Samples)} This is the classic CEM used in the original PETS algorithm~\cite{pets}, configured to evaluate $N$ total sampled rollouts. Recall that CEM samples $n$ action sequences each iteration, and iteratively updates the sampling distribution for $m$ total iterations, with a total of $N = nm$ sampled rollouts per step. For $N = 500$ and $N = 5000$, we run CEM for $m = 5$ and $m = 50$ iterations, respectively, with $n = 100$ samples per iteration. For $N = 50$, we run CEM for $m=5$ iterations, with $n=10$ samples per iteration. 
    \item \textbf{Baseline Gradient Optimizer} This is the classic first-order optimization method without any sampling. A random initialization action sequence is selected, and gradient descent is performed, using the same line search method as described in section \ref{gradient_descent}.  
    \item \textbf{CEM-GD} Our proposed planner, as described in Algorithm \ref{alg:main_alg}. We combine CEM and the first-order optimization method, by using CEM to select initial action sequences and optimizing using gradient descent. For all experiments, we use $N_{init} = 15000$ and $N_{r} = 50$.
\end{itemize}

For the baseline gradient optimizer and \alg, we use $k = 1$ (number of initialization action sequences for optimization), $G = 10$ (number of gradient update steps), $\eta_{init} = 0.01$ (initial step size for gradient descent), $\rho = 0.67$ (step size decay), and $J = 8$ (maximum line search trials). For all experiments, the planning horizon length $T$ is 45. The dynamics model is a 3 layer multi-layer perceptron with 200 neurons per layer and SiLU activation functions \citep{silu}, pretrained with 500 random trajectory rollouts in the environment. 

\subsection{Performance}
We compare performances of the proposed CEM-GD planner, the baseline CEM method and the baseline gradient optimizer in four control tasks. The Pusher environment is the one used by \citeauthor{pets}, and the Humanoid environment is the truncated-state version used by \citeauthor{MBPO}. 
As shown in Figure \ref{fig:results}, our proposed planner performs as well as or better than CEM with 100 times less samples. For the Half-Cheetah and Humanoid Standup tasks (Figure \ref{fig:results} (c) and (d)), we see that the performance of CEM is heavily affected by the number of samples used. If we decrease the number of samples from $5000$ to $500$, the performance of CEM deteriorates significantly. However, with only $50$ samples, \alg is able to achieve better performance than CEM with $5000$ samples, demonstrating the improvement in sample efficiency by leveraging gradients. In the Cartpole Swingup and Pusher tasks (Figure \ref{fig:results} (a) and (b)), the performance is saturated after using around $500$ samples in CEM. In the case of Cartpole Swingup, this is likely due to the action space having a dimensionality of one. However, our planner is able to achieve a similar level of performance with only $50$ samples. 

In addition, we compare against the baseline gradient optimizer (the grey lines in Figure \ref{fig:results}), which does not use any sampling. Because a pure first-order optimizer can be susceptible to poor local optima, its performances varies wildly across different environments. However, our method which combines zeroth-order and first-order methods, performs consistently in all environments.
\begin{figure}[t]
	\centering
	\subfigure[]{\label{fig:swingup}\includegraphics[width=0.45\columnwidth]{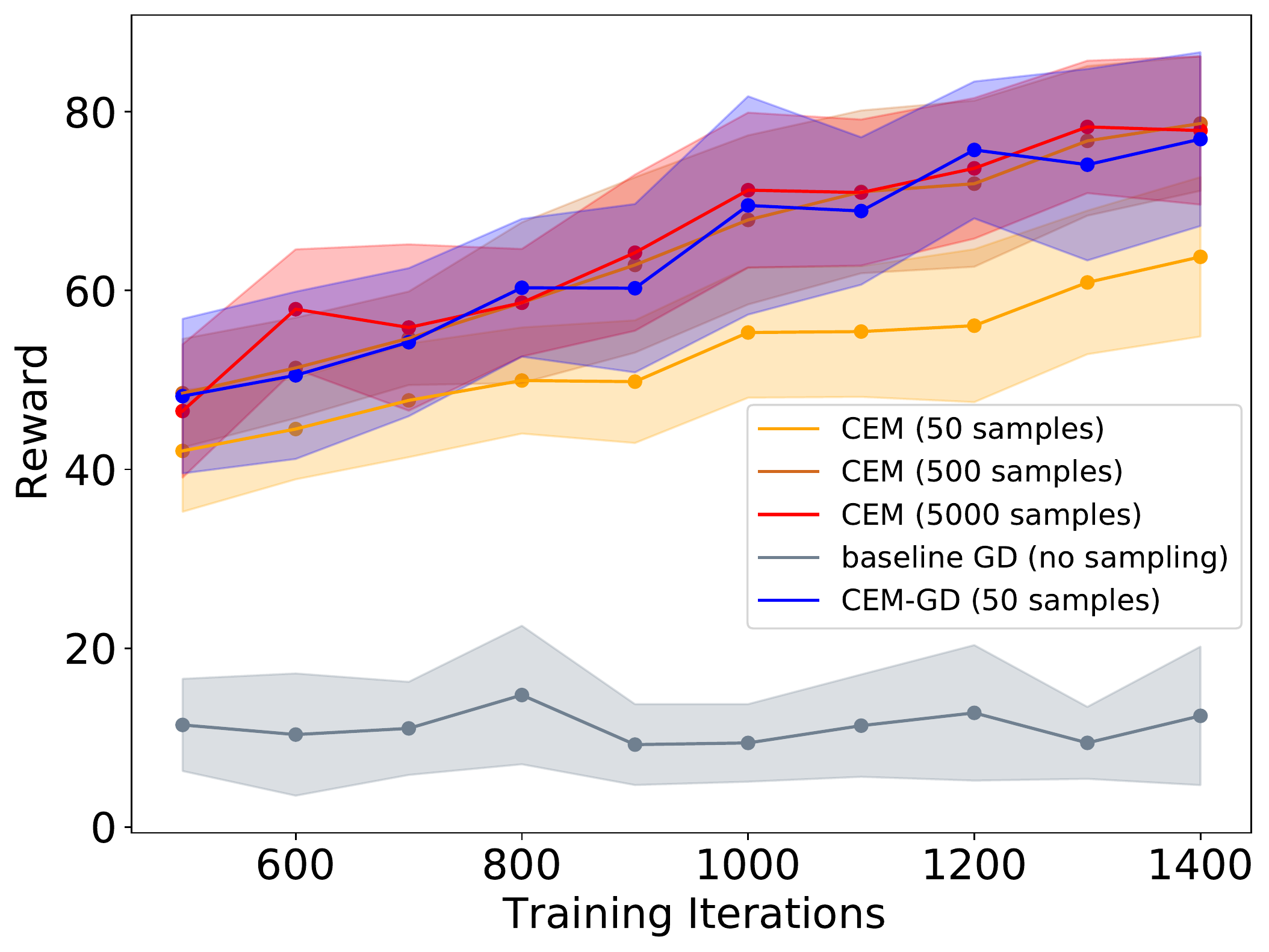}} 
	\subfigure[]{\label{fig:pusher}\includegraphics[width=0.45\columnwidth]{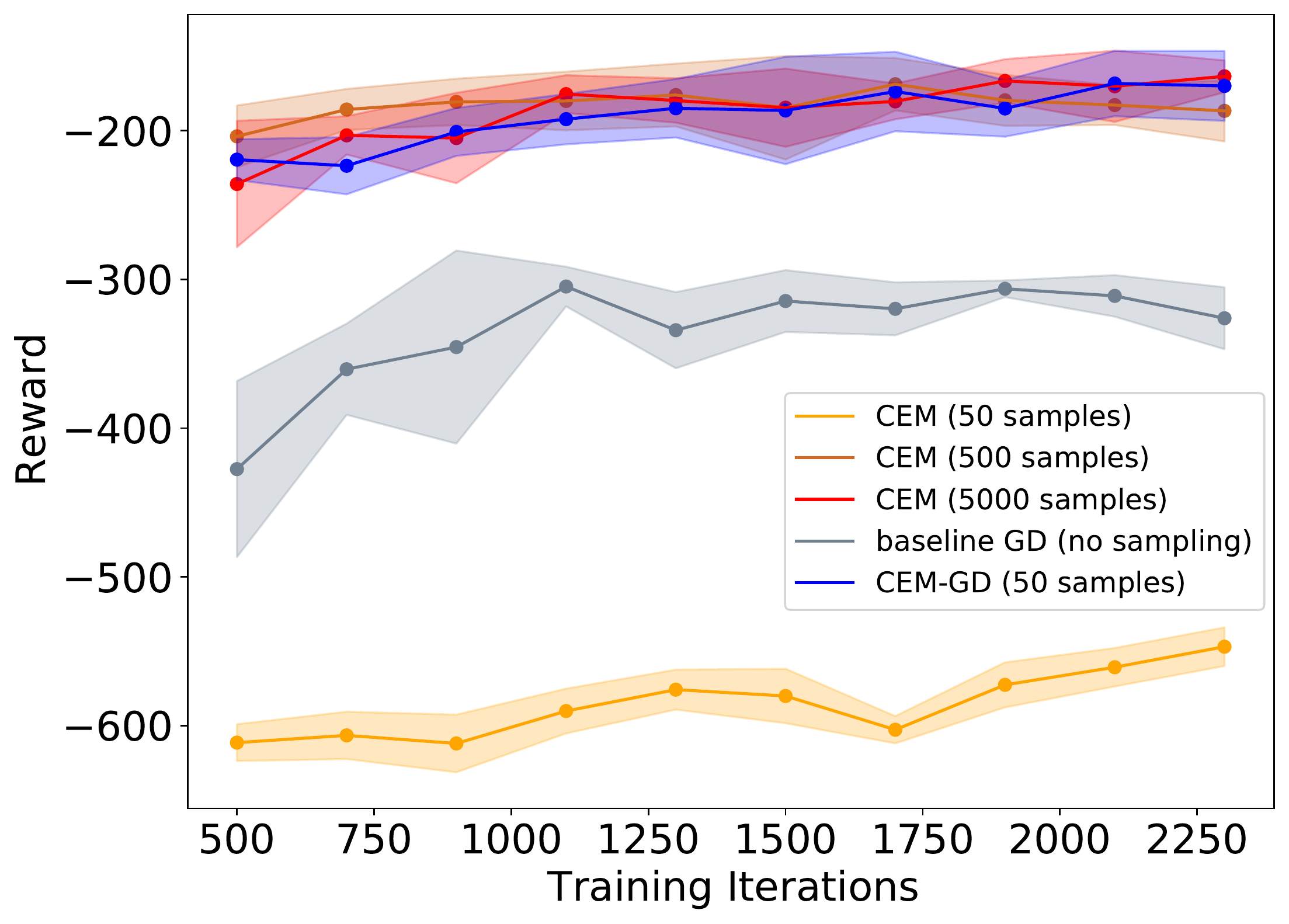}} 
	\subfigure[]{\label{fig:cheetah}\includegraphics[width=0.45\columnwidth]{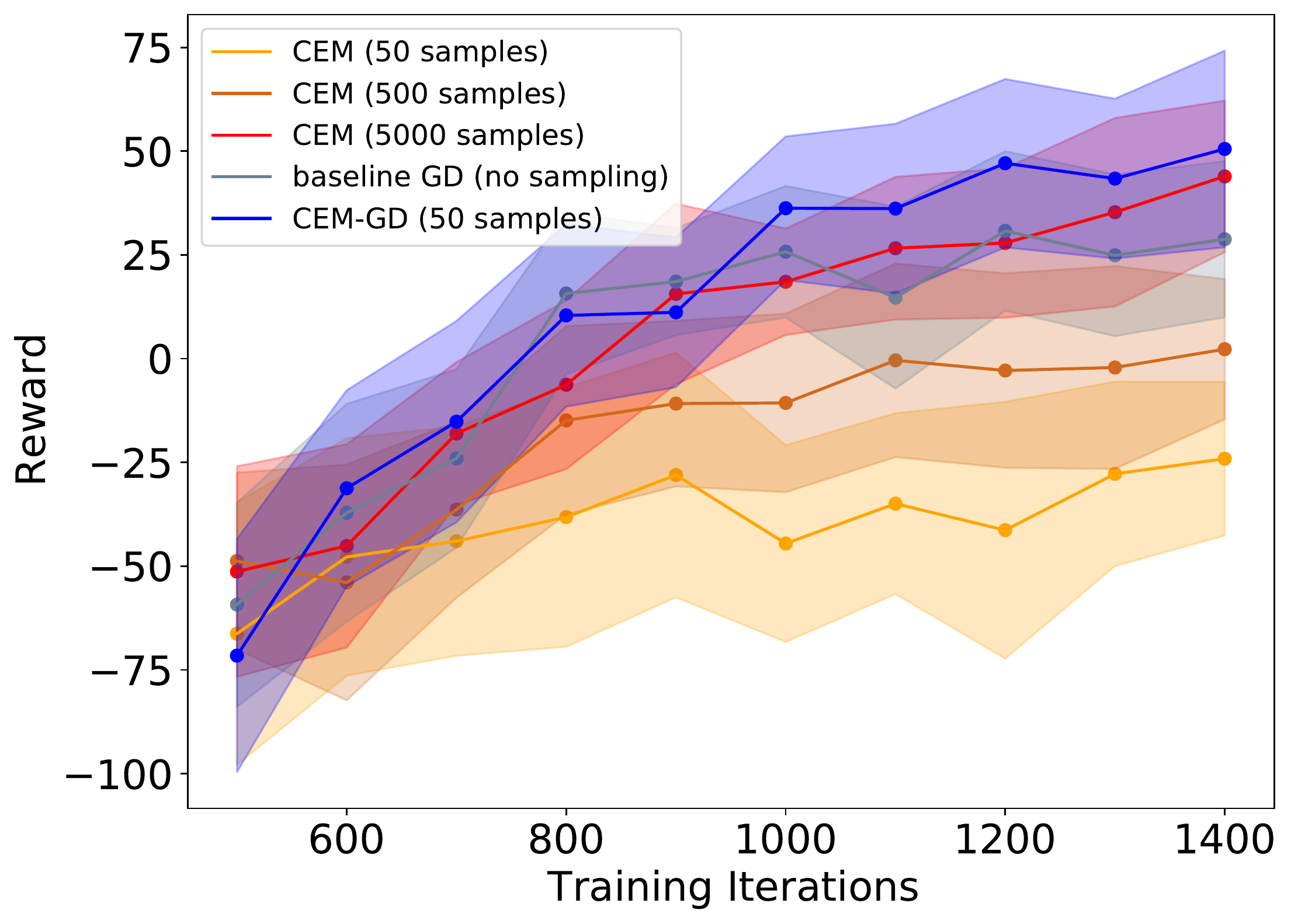}} 
	\subfigure[]{\label{fig:humanoid}\includegraphics[width=0.45\columnwidth]{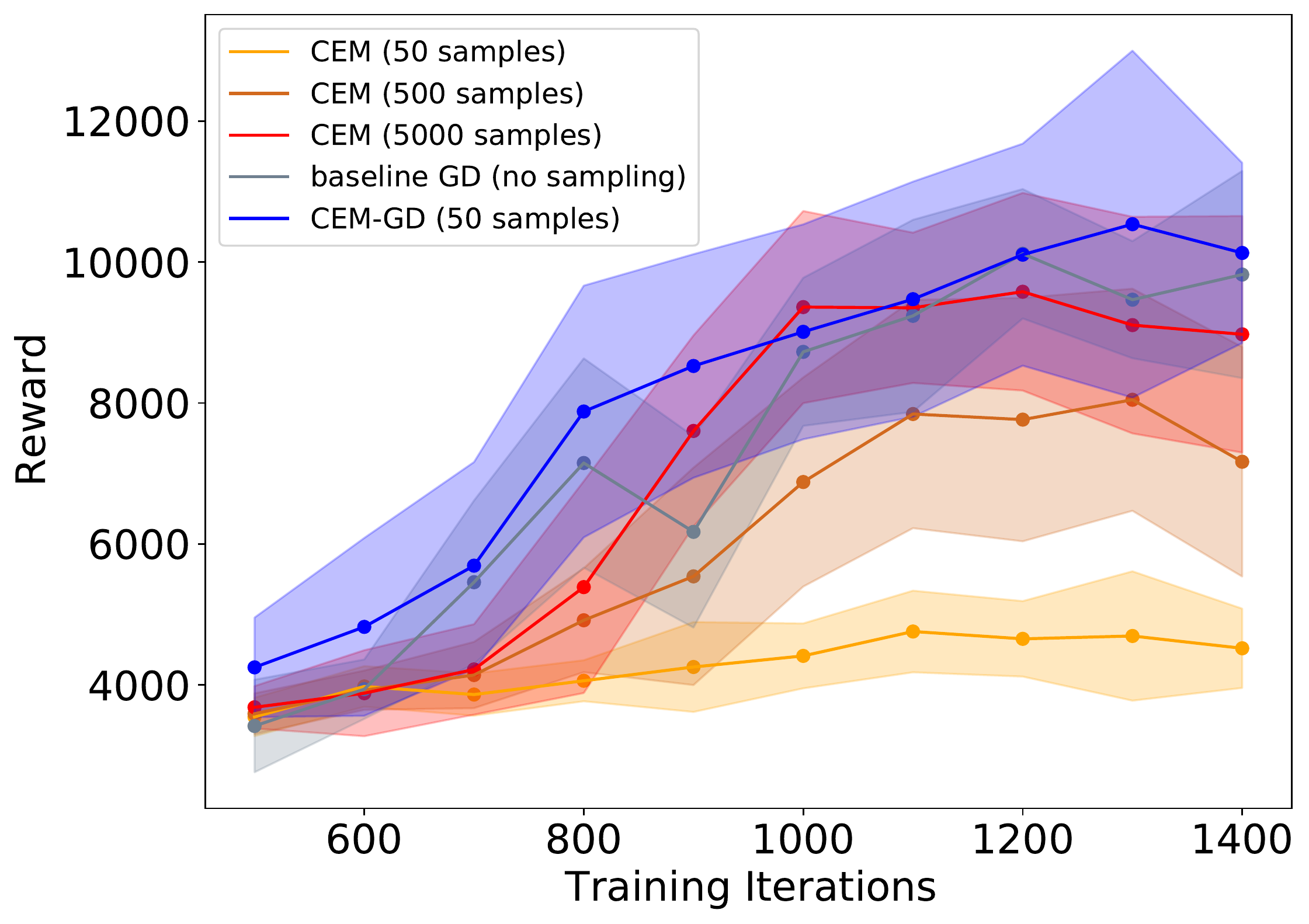}} 
	\caption{Comparison of different planners on the (a) Cartpole-Swingup, (b) Pusher, (c) Half-Cheetah, (d) Humanoid Standup. The mean and standard deviation of the rewards across 20 experiments is plotted.}
	\label{fig:results}
\end{figure}

\subsection{Sample Efficiency}
\begin{figure}[t]
	\centering
	\includegraphics[width=0.6\columnwidth]{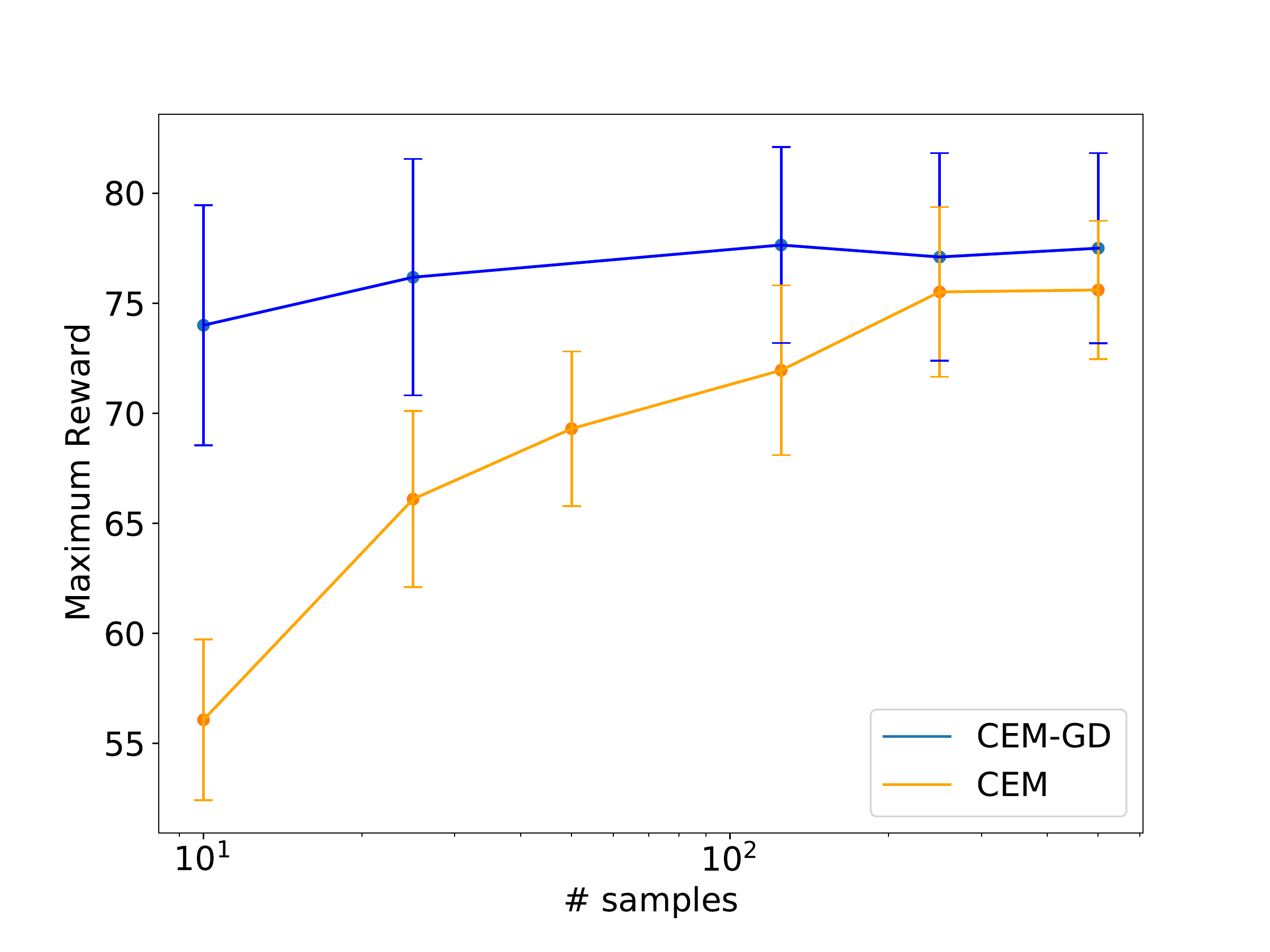}
	\caption{\label{fig:sample_efficiency} Comparison of the sample efficiency of CEM with \alg on the Cartpole Swingup Task. CEM greatly suffers from reducing the number of sampled rollouts taken, but not our planner which leverages gradients. For \alg, the x-axis represents $N_{r}$.}
	\label{fig:results}
\end{figure}

We further compare the sample efficiency of CEM and our method in Fig \ref{fig:sample_efficiency}. The y-axis is the maximum reward obtained after 1200 training iterations, averaged over 20 experiments. The x-axis is the total number of rollouts sampled during CEM; for our planner, this is $N_{r}$. 
As shown in Figure \ref{fig:sample_efficiency}, for standard CEM planner, decreasing the number of samples significantly reduces performance, while the number of samples has little effect on the performance for our gradient based planner. 

From the analysis shown in Figure \ref{fig:sample_efficiency} and from \cite{gradcem}, as the dimensionality of the problem increases, we would expect that CEM performs relatively worse given a fixed budget of samples. Indeed, in these experiments, we see that for the Half-Cheetah and Humanoid environments, which have action dimensions of $6$ and $17$, respectively, increasing the number of samples from $500$ to $5000$ improves the performance of CEM. However, our gradient based planner works across all environments with a budget of $N_r = 50$ samples, despite differences in dimensionality, which demonstrated preferrable sample efficiency.

\subsection{Computation Time and Memory Usage}
We show the computation time for different methods in Table \ref{table:compute} and the peak memory usage in Table \ref{table:memory}. \alg is faster than CEM with $5000$ samples, and uses less memory, though it equals or exceeds its performance in the experiments. 
\begin{table}[htbp]
\centering
\caption{\label{table:compute} Average runtime for the various planners tested, in seconds. }
\vskip 0.3cm
\centering
\renewcommand{\arraystretch}{1.2}
\resizebox{0.55\linewidth}{!}{%
\begin{tabular}{c c c c c} 
 \hline
 \hline
 Planner & Cartpole & Half-Cheetah & Humanoid & Pusher \\ [0.5ex] 
 \hline
 CEM 50 & 0.36 & 0.36 & 0.39 & 0.36 \\
 CEM 500 & 0.39  & 0.39 & 0.41 & 0.38 \\ 
 CEM 5000 & 3.35 & 3.38 & 3.44 & 3.33 \\
 \alg & 2.18 & 2.63 & 2.59 & 2.58 \\
 \hline
\end{tabular}}
\end{table}

\begin{table}[htbp]
\centering
\caption{\label{table:memory} Peak memory usage (MB) of \alg vs CEM. }
\vskip 0.3cm
\centering
\renewcommand{\arraystretch}{1.2}
\resizebox{0.55\linewidth}{!}{%
\begin{tabular}{c c c c c} 
 \hline
 \hline
 Planner & Cartpole & Half-Cheetah & Humanoid & Pusher \\ [0.5ex] 
 \hline
 CEM 50 & 7.2 & 8.1 & 9.9 & 8.3 \\
 CEM 500 & 12.9 & 14.3 & 17.3 & 14.4 \\ 
 CEM 5000 & 12.9 & 14.3 & 17.3 & 14.4 \\
 \alg & 11.4 & 12.2 & 14.3 & 12.4 \\
 \hline
\end{tabular}}
\end{table}

As the dimensionality of the optimal control problem increases, CEM requires more samples to perform well. For example, we see that CEM with 5000 samples has a significant performance increase over 500 samples for Half-Cheetah, but not for the Cartpole Swingup task. As Table \ref{table:compute} demonstrates, increasing the number of samples for CEM is computationally expensive. While \alg incurs additional computation time due to backpropagation to obtain the gradients, this computation time is fixed, as we apply gradient updates to only $k$ initialization points where $k$ is fixed. Thus, our planner scales much better asymptotically for higher dimensional planning problems. Similarly, due to the increased sample efficiency of \alg, it needs to sample fewer rollouts during the CEM step, reducing the memory usage.  

\section{Related Work}

Our work is in the context of model-based reinforcement learning and model predictive control, where an agent learns a dynamics model of the environment, and uses rolls out trajectories using the model to determine an optimal action to execute. Specifically, we designed our planner in the context of PETS \cite{pets}, where a probabilistic ensemble is used as the dynamics model and which has achieved state-of-the-art results. When performing our experiments, we compare our CEM-GD planner, baseline gradient planner and standard planner in PETS. 

The Cross-Entropy Method \cite{CEM} is an effective general purpose zeroth-order optimizer that has been used for a wide variety of applications. Most state of the art MBRL methods use CEM to optimize over a set of plans to select an optimal control action \cite{pets, Hafner, poplin}. In particular, \cite{poplin} uses a policy network and uses CEM to optimize over the parameter space of the policy. However, MBRL methods using CEM typically do not also investigate the sample efficiency of CEM and the scalability to high planning dimensionality, especially when compared to a gradient based planner.
There have been several efforts to improve sample efficiency and performance for CEM. For example, \citeauthor{icem} proposed to more efficiently sample the action space in various frequency spectra but it requires prior knowledge of the environment or parameter tuning for the particular frequency spectrum. A recent work \citet{gradcem} proposed to perform gradient updates within CEM to improve its performance, but it performs gradient updates to each sample thus still suffering poor scalability.




\section{Conclusions and Discussions}
To summarize, we propose CEM-GD, a planner that combines zeroth order methods like CEM with first order gradient-based optimization in order to achieve better scalability for high dimensional planning problems. We demonstrate that as the problem dimension grows, more samples are needed for CEM to converge, which is computationally expensive. However, by leveraging gradients, CEM-GD obtains better sample complexity, reduces computation time and memory usage. 

While we run our experiments in the context of MBRL with MPC, specifically PETS, it is important to note that we focus only on proposing a planner, not a full end-to-end reinforcement learning algorithm. In particular, our planner does not make any assumptions about the dynamics model, which likely has modelling error. While PETS uses a probabilistic ensemble that includes model uncertainty \cite{pets}, this uncertainty is not utilized during the gradient descent step of our planner. That is, when rolling out a trajectory to compute the gradients, our planner uses only the mean prediction of the ensemble. To design a full end-to-end MBRL algorithm, model uncertainty would have to be integrated into the planner. One possibility would be to add noise to the planner's selected actions based on model uncertainty. 

\bibliography{sample}

\begin{thebibliography}{18}
\providecommand{\natexlab}[1]{#1}
\providecommand{\url}[1]{\texttt{#1}}
\expandafter\ifx\csname urlstyle\endcsname\relax
  \providecommand{\doi}[1]{doi: #1}\else
  \providecommand{\doi}{doi: \begingroup \urlstyle{rm}\Url}\fi

\bibitem[Bharadhwaj et~al.(2020)Bharadhwaj, Xie, and Shkurti]{gradcem}
Homanga Bharadhwaj, Kevin Xie, and Florian Shkurti.
\newblock Model-predictive control via cross-entropy and gradient-based
  optimization.
\newblock In \emph{Proceedings of the 2nd Annual Conference on Learning for
  Dynamics and Control}, 2020.
\newblock URL \url{http://proceedings.mlr.press/v120/bharadhwaj20a.html}.

\bibitem[Botev et~al.(2013)Botev, Kroese, Rubinstein, and L’Ecuyer]{CEM}
Zdravko~I. Botev, Dirk~P. Kroese, Reuven~Y. Rubinstein, and Pierre L’Ecuyer.
\newblock Chapter 3 - the cross-entropy method for optimization.
\newblock In \emph{Handbook of Statistics}, volume~31 of \emph{Handbook of
  Statistics}, pages 35--59. Elsevier, 2013.
\newblock \doi{https://doi.org/10.1016/B978-0-444-53859-8.00003-5}.
\newblock URL
  \url{https://www.sciencedirect.com/science/article/pii/B9780444538598000035}.

\bibitem[Brockman et~al.(2016)Brockman, Cheung, Pettersson, Schneider,
  Schulman, Tang, and Zaremba]{gym}
Greg Brockman, Vicki Cheung, Ludwig Pettersson, Jonas Schneider, John Schulman,
  Jie Tang, and Wojciech Zaremba.
\newblock {OpenAI Gym}, 2016.
\newblock arXiv:1606.01540.

\bibitem[Chua et~al.(2018)Chua, Calandra, McAllister, and Levine]{pets}
Kurtland Chua, Roberto Calandra, Rowan McAllister, and Sergey Levine.
\newblock Deep reinforcement learning in a handful of trials using
  probabilistic dynamics models.
\newblock In \emph{Advances in Neural Information Processing Systems 31}, pages
  4759--4770, 2018.
\newblock URL
  \url{https://proceedings.neurips.cc/paper/2018/hash/3de568f8597b94bda53149c7d7f5958c-Abstract.html}.

\bibitem[Elfwing et~al.(2018)Elfwing, Uchibe, and Doya]{silu}
Stefan Elfwing, Eiji Uchibe, and Kenji Doya.
\newblock Sigmoid-weighted linear units for neural network function
  approximation in reinforcement learning.
\newblock \emph{Neural Networks}, 107:\penalty0 3--11, 2018.
\newblock \doi{10.1016/j.neunet.2017.12.012}.
\newblock URL \url{https://doi.org/10.1016/j.neunet.2017.12.012}.

\bibitem[Hafner et~al.(2019)Hafner, Lillicrap, Fischer, Villegas, Ha, Lee, and
  Davidson]{Hafner}
Danijar Hafner, Timothy~P. Lillicrap, Ian Fischer, Ruben Villegas, David Ha,
  Honglak Lee, and James Davidson.
\newblock Learning latent dynamics for planning from pixels.
\newblock In \emph{Proceedings of the 36th International Conference on Machine
  Learning}, pages 2555--2565, 2019.
\newblock URL \url{http://proceedings.mlr.press/v97/hafner19a.html}.

\bibitem[Hauser(2007)]{hauser2007line}
Raphael Hauser.
\newblock Line search methods for unconstrained optimisation.
\newblock \emph{Lecture 8, Numerical Linear Algebra and Optimisation Oxford
  University Computing Laboratory}, 2007.

\bibitem[Janner et~al.(2019)Janner, Fu, Zhang, and Levine]{MBPO}
Michael Janner, Justin Fu, Marvin Zhang, and Sergey Levine.
\newblock When to trust your model: Model-based policy optimization.
\newblock In \emph{Advances in Neural Information Processing Systems 32}, pages
  12498--12509, 2019.
\newblock URL
  \url{https://proceedings.neurips.cc/paper/2019/hash/5faf461eff3099671ad63c6f3f094f7f-Abstract.html}.

\bibitem[Lale et~al.(2021)Lale, Azizzadenesheli, Hassibi, and
  Anandkumar]{lale2021model}
Sahin Lale, Kamyar Azizzadenesheli, Babak Hassibi, and Anima Anandkumar.
\newblock Model learning predictive control in nonlinear dynamical systems.
\newblock 2021.

\bibitem[Nagabandi et~al.(2018)Nagabandi, Kahn, Fearing, and
  Levine]{nagabandi2018neural}
Anusha Nagabandi, Gregory Kahn, Ronald~S Fearing, and Sergey Levine.
\newblock Neural network dynamics for model-based deep reinforcement learning
  with model-free fine-tuning.
\newblock In \emph{2018 IEEE International Conference on Robotics and
  Automation (ICRA)}, pages 7559--7566. IEEE, 2018.

\bibitem[Okada and Taniguchi(2020)]{okada2020variational}
Masashi Okada and Tadahiro Taniguchi.
\newblock Variational inference mpc for bayesian model-based reinforcement
  learning.
\newblock In \emph{Conference on Robot Learning}, pages 258--272. PMLR, 2020.

\bibitem[Pinneri et~al.(2020)Pinneri, Sawant, Blaes, Achterhold,
  St{\"{u}}ckler, Rolinek, and Martius]{icem}
Cristina Pinneri, Shambhuraj Sawant, Sebastian Blaes, Jan Achterhold,
  J{\"{o}}rg St{\"{u}}ckler, Michal Rolinek, and Georg Martius.
\newblock Sample-efficient cross-entropy method for real-time planning.
\newblock In \emph{4th Conference on Robot Learning}, pages 1049--1065, 2020.
\newblock URL \url{https://proceedings.mlr.press/v155/pinneri21a.html}.

\bibitem[Rubinstein(1997)]{rubinstein1997optimization}
Reuven~Y Rubinstein.
\newblock Optimization of computer simulation models with rare events.
\newblock \emph{European Journal of Operational Research}, 99\penalty0
  (1):\penalty0 89--112, 1997.

\bibitem[Srinivas et~al.(2018)Srinivas, Jabri, Abbeel, Levine, and
  Finn]{srinivas2018universal}
Aravind Srinivas, Allan Jabri, Pieter Abbeel, Sergey Levine, and Chelsea Finn.
\newblock Universal planning networks: Learning generalizable representations
  for visuomotor control.
\newblock In \emph{International Conference on Machine Learning}, pages
  4732--4741. PMLR, 2018.

\bibitem[Todorov et~al.(2012)Todorov, Erez, and Tassa]{mujoco}
Emanuel Todorov, Tom Erez, and Yuval Tassa.
\newblock Mujoco: {A} physics engine for model-based control.
\newblock In \emph{International Conference on Intelligent Robots and Systems},
  2012.

\bibitem[Von~Stryk and Bulirsch(1992)]{von1992direct}
Oskar Von~Stryk and Roland Bulirsch.
\newblock Direct and indirect methods for trajectory optimization.
\newblock \emph{Annals of operations research}, 37\penalty0 (1):\penalty0
  357--373, 1992.

\bibitem[Wang and Ba(2020)]{poplin}
Tingwu Wang and Jimmy Ba.
\newblock Exploring model-based planning with policy networks.
\newblock In \emph{8th International Conference on Learning Representations},
  2020.
\newblock URL \url{https://openreview.net/forum?id=H1exf64KwH}.

\bibitem[Williams et~al.(2017)Williams, Aldrich, and
  Theodorou]{williams2017model}
Grady Williams, Andrew Aldrich, and Evangelos~A Theodorou.
\newblock Model predictive path integral control: From theory to parallel
  computation.
\newblock \emph{Journal of Guidance, Control, and Dynamics}, 40\penalty0
  (2):\penalty0 344--357, 2017.

\end{thebibliography}

\end{document}